\documentclass{article}
\usepackage{spconf}
\usepackage{amsmath}
\usepackage{graphicx}
\usepackage{epstopdf}
\usepackage{multirow}
\usepackage{color}
\usepackage{url}
\usepackage{pifont}
\newcommand{\cmark}{\ding{51}}%
\newcommand{\xmark}{\ding{53}}%

\title{UTD-CRSS Systems for 2016 NIST Speaker Recognition Evaluation}
\name{Chunlei Zhang, Fahimeh Bahmaninezhad, Shivesh Ranjan, Chengzhu Yu, Navid Shokouhi, John H.L. Hansen}

\address{Center for Robust Speech Systems (CRSS), Erik Jonsson School of Engineering, \\
	 University of Texas at Dallas, Richardson, Texas, U.S.A. \\
	 \small \tt \{chunlei.zhang,fahimeh.bahmaninezhad,john.hansen\}@utdallas.edu}
\begin{document}
\ninept
\maketitle
\begin{abstract}
This document briefly describes the systems submitted by the Center for Robust Speech Systems (CRSS) from The University of Texas at Dallas (UTD) to the 2016 National Institute of Standards and Technology (NIST) Speaker Recognition Evaluation (SRE). We developed several UBM and DNN i-Vector based speaker recognition systems with different data sets and feature representations. Given that the emphasis of the NIST SRE 2016 is on language mismatch between training and enrollment/test data, so-called domain mismatch, in our system development we focused on: (1) using unlabeled in-domain data for centralizing data to alleviate the domain mismatch problem, (2) finding the best data set for training LDA/PLDA, (3) using newly proposed dimension reduction technique incorporating unlabeled in-domain data before PLDA training, (4) unsupervised speaker clustering of unlabeled data and using them alone or with previous SREs for PLDA training, (5) score calibration using only unlabeled data and combination of unlabeled and development (Dev) data as separate experiments.

\end{abstract}
%
%
\section{INTRODUCTION}

\label{sec:intro}
The main task for NIST's speaker recognition evaluations is speaker detection, i.e., to determine whether a specified target speaker is speaking during a given segment of speech or not. Compared with previous SRE challenges, there are some differences: (1) target speaker data will not be distributed in advance like in SRE12, (2) fixed condition is introduced, (3) more duration variability is introduced in the test data, (4) language mismatch between training (mainly English) and enrollment/test (non-English) data. All these new traits make this SRE very challenging, especially with limited labeled data in the fixed condition \cite{2016nist}. 

This report introduces how CRSS systems address the problem. The whole report is organized as follows: Sec.\ref{sec:baseline} describes several baseline systems focusing on the front-end level overview, including both data sets and feature representations, Sec.\ref{sec:framework} introduces several core techniques that we used in SRE16, including speaker clustering for unlabeled training data, discriminant analysis via support vectors (SVDA) to reduce dimension and compensate domain mismatch, unlabeled data PLDA, calibration and fusion strategies etc. Sec.\ref{sec:exp} and Sec.\ref{submissions} details the configuration of every CRSS sub-system and the formation of CRSS final Eval submissions to NIST. Sec.\ref{performance} shows the CRSS sub-system performance on Dev data. We also briefly introduce the CPU/GPU hardware systems that we used for NIST SRE 2016 in Sec.\ref{Computational}. 

\section{CRSS baselines}

\label{sec:baseline}
We developed 4 baseline systems in this SRE, all of them are i-Vector based systems but with different acoustic modeling, i.e., UBM or different DNN models \cite{lei2014novel,sadjadi2016ibm}. For back-end, we mainly use LDA/SVDA to reduce the dimension of i-Vectors and PLDA to calculate likelihood scores.

Table \ref{tab:data_statistics} summarized the number of speakers, speech segments used for training UBM, total variability matrix (TV), LDA/SVDA and PLDA models as well as the statistics for the Dev set provided by NIST for the system development purposes.

      \begin{table*}[th]
        \caption{\label{tab:data_statistics} {\it Statistics of training data used for modeling DNN, UBM, TV, LDA/SVDA/PLDA, and also Dev set enrollment and trials data.}}
        \vspace{2mm}
        \centerline{
		\begin{tabular}{ |c||c|c||c|c||c|c||c|c||c|c| }
  			\hline
  			\multicolumn{1}{|c||}{Sub-system} &
  			\multicolumn{2}{|c||}{DNN} &
  			\multicolumn{2}{|c||}{UBM/TV} &
  			\multicolumn{2}{|c||}{LDA/SVDA/PLDA} &
  			\multicolumn{2}{|c||}{Enrollment} &
  			\multicolumn{2}{|c|}{Trials}\\
  			  & Spkrs & Segments & Spkrs & Segments & Spkrs & Segments & Spkrs & Segments & Target & nonTarget \\
  			\hline
  			\hline
  			CRSS1 & - & - & 38110 & 89860 & 2921 & 37037 & \multirow{4}{*}{80} & \multirow{4}{*}{120} & \multirow{4}{*}{4828} & \multirow{4}{*}{19312} \\
  			CRSS2 & 4878 & 4878 & 5767 & 57517 & 2921 & 37037 &  &  &  &\\
  			CRSS3 & - & - & 5756 & 57273 & 3794 & 36410  &  &  &  &\\
  			CRSS4 & 1239 & 1239 & 5756 & 57273 & 3794 & 36410 &  &  &  &\\
  			\hline
		\end{tabular}
		}
      \end{table*}

\subsection{CRSS1: UBM i-Vector}
This system is mainly modified version of Kaldi (sre10/v1). 60 dimensional feature vectors for each frame is adopted here including 20 dimensional MFCC features appended with $\Delta+\Delta\Delta$. Unvoiced parts of the utterances are removed with energy based voice activity detection (VAD). For training 2048-mixture UBM and total variability (TV) matrix, SRE2004, 2005, 2006, 2008, telephone data of SRE 2010, Switchboard II phase 2,3 and Switchboard Cellular Part1 and Part2 (SWB) and Fisher English are used. Next, 600 dimensional i-Vectors are extracted and their dimensions are reduced to 580 with LDA. For training LDA/PLDA, only SRE 04-08 are used; in addition, speakers who have less than 4 utterances is filtered out. Also, unsupervised speaker clustering is performed (see Sec.~\ref{subsec:speaker_cluster} for the details of speaker clustering), 75 speaker clusters for unlabeled minor data and 300 for unlabeled major data are generated. These clustered in-domain data are used separately to train PLDA and also for calibration. Before PLDA scoring, mean subtraction is also applied. For SRE16 development (Dev) trials, the mean i-Vector is generated using only unlabeled minor data, while for SRE16 evaluation (Eval), the mean is from unlabeled major data.

\subsection{CRSS2: SWB DNN i-Vector}
We developed a DNN i-Vector system based on Kaldi (swbd/s5 \& sre10/v2). The DNN acoustic model is used to generate the soft alignments for i-Vector extraction. The DNN architecture has 6 fully connected hidden layers with 1024 nodes for each layer. Cross-entropy objective function is employed to estimate posterior probabilities of 3178 senones. The ASR corpus which we used for training DNN acoustic model is Switchboard. 11-frame context of 39 dimensional ($\Delta+\Delta\Delta$ ) MFCC feature is projected into 40 dimensional using fMLLR transform for each utterance, which relies on a GMM-HMM decoding alignment.

The reason we apply fMLLR feature here is that, by speaker normalization, we expect to acquire more accurate phonetic alignment in the following TV matrix training, see more details in \cite{sadjadi2016ibm}. After i-Vector extraction, we apply similar strategies for back-end such as LDA and PLDA, briefly described in the above section (as CRSS1).

\subsection{CRSS3: UBM i-Vector}

An alternative UBM i-Vector system also adopted from Kaldi (sre10/v1). In this system, feature vectors contain 20 MFCCs appended with ($\Delta+\Delta\Delta$) coefficients. The window length and shift size are 25-ms and 10-ms, respectively. In addition, we did cepstral mean normalization using 3-sec sliding window. Next, non-speech frames are discarded using energy-based voice activity detection. 2048-mixture full covariance UBM and total variability matrix have been trained using data collected from SRE2004, 2005, 2006, 2008 and Switchboard II phase 2,3 and Switchboard Cellular Part1 and Part2. At the back-end, after extracting i-Vectors, the global mean calculated from minor and major unlabeled data is subtracted from all i-Vectors. Next, i-Vectors are length-normalized and their dimension are reduced from 600 to 400 using LDA/SVDA. In some developed systems based on CRSS3 configuration, we used both LDA and SVDA; more specifically, first SVDA reduces the dimension from 600 to 500 and then LDA is used to reduce the dimension to 400. Again, i-Vectors are length-normalized. Finally, trial-based mean subtraction is used (the participant i-Vectors in a trial are averaged and the value is subtracted from both i-Vectors) and scores are calculated using PLDA.
The front end is trained with SWB and SRE04-08; however, the back-end only uses SRE04-08 and unlabeled training data. For back-end, mostly MSR \cite{sadjadi2013msr} toolkit has been adopted.

\subsection{CRSS4: Fisher English DNN i-Vector}
The last baseline is a DNN i-Vector system using Kaldi (sre10/v2) that is based on the multisplice time delay DNN (TDNN) \cite{snyder2015time}. TDNN is trained with only a small portion of Fisher English data (1239 utterances). The feature vectors contain 40 dimensional f-bank features. TDNN has six layers; the hidden layers have an input dimension of 350 and an output dimension 3500. The softmax output layer computes posteriors for 3859 triphone states. More details on the TDNN structure and training procedure are provided in \cite{snyder2015time}. After TDNN training, 20 MFCCs appended with ($\Delta+\Delta\Delta$) coefficients (overall 60 MFCCs) are employed for training TV matrix. Next, 600-dim i-Vectors are extracted.

After i-Vector extraction, we apply similar strategies for back-end such as LDA/SVDA and PLDA, briefly described in the above section (as CRSS3).

\section{Core components in system development}
\label{sec:framework}
In the fixed condition of SRE 2016, we have a huge amount of out-of-domain data, i.e., previous SREs, SWB, Fisher English etc. Only a small in-domain data is available (without speaker labels), which make existing techniques very difficult to work with this so-called domain mismatch. In SRE 2016, NIST provided unlabeled training data, which contains two subsets, i.e., unlabeled minor and unlabeled major. The unlabeled minor data set has 200 utterances, while the major set contains 2272 utterances. The minor set has two languages for the purpose of system development, while the major set contains two different languages corresponding to the final evaluation.

In order to address this problem, several techniques are proposed in this evaluation.

\subsection{Speaker clustering of unlabeled data} 
\label{subsec:speaker_cluster}

For compensating the domain mismatch, the use of unlabeled data becomes very important. There are several stages where we can use the unlabeled data, such as, LDA/PLDA training and calibration. First, it is very intuitive to do a speaker clustering of the unlabeled data, and then generate an ``estimated'' speaker label for each utterance, similar with the method that we used in 2015 NIST LRE i-Vector challenge\cite{yu2016utd}. With these labels, we incorporate the in-domain information from unlabeled data to train LDA and PLDA. In fact, in the experiment, this simple operation improved the LDA/PLDA baseline performance for development set.

In practice, we train a gender identification using previous SRE data before speaker clustering, and then apply a simple K-means algorithm over gender dependent subsets, finally, we pool these two subsets together. In the experiment, we found this can provide more accurate speaker clustering and more benefits to the following LDA and PLDA training.

\begin{table*}[!th]
	\centering
	\caption{\label{Table1} {\it Description of the CRSS sub-systems. Speaker clustering means using the unlabeled data to do clustering and using estimated labels for LDA/PLDA training. The systems that have SVDA, for training SVDA they use LDA data in addition to Minor and Major data. For sub-systems 3 and 4, for Dev set Minor data are used for training PLDA; however, for Eval set Major data are used.}}
	\vspace{2mm}
	\resizebox{14cm}{!}{
	\begin{tabular}{|c|c|c|c|c|c|c|c|c|}
		\hline		
		 \begin{tabular}[c]{@{}c@{}}Sub\\system\end{tabular} &{i-Vector}& \begin{tabular}[c]{@{}c@{}}LDA/SVDA data\end{tabular}  & 
		 \begin{tabular}[c]{@{}c@{}}PLDA data\end{tabular} & {SVDA}  &LDA& \begin{tabular}[c]{@{}c@{}}Speaker \\ Clustering\end{tabular}   & {Filtering}\\
		\hline\hline
		 1 &CRSS 4  & SRE 04-08 & SRE 04-08 & \xmark &\cmark& \xmark& \xmark\\
		\hline
		 2 &CRSS 4 &SRE 04-08/ + Minor, Major & SRE 04-08  &\cmark &\cmark& \xmark & \xmark \\
		\hline
	     3 & CRSS 2  &SRE 04-08, Minor, Major & SRE 04-08, Minor or Major  &\xmark &\cmark& \cmark &\cmark \\
		\hline
		 4 &CRSS 1  &SRE 04-08, Minor, Major & SRE 04-08, Minor or Major  &\xmark &\cmark& \cmark & \cmark  \\
		\hline
		 5 &CRSS 3  &SRE 04-08 & SRE 04-08 &\xmark &\cmark& \xmark& \xmark  \\
		\hline
		 6 &CRSS 3  &SRE 04-08/+ Minor, Major & SRE 04-08  &\cmark &\cmark& \xmark& \xmark \\
		\hline
	     7&CRSS 3  & SRE 04-08/+ Minor, Major & SRE 04-08  &\cmark &\xmark& \xmark&  \xmark \\
		\hline
	\end{tabular}}
\end{table*}

\subsection{Discriminant analysis via support vectors (SVDA)} 
\label{subsec:svda}

Discriminant analysis via support vectors (SVDA) is a variation of LDA that only uses support vectors to calculate the between and within class covariance matrices. In contrast to LDA, SVDA captures the boundary of classes, and performs well for small sample size problem (i.e. when the dimensionality is greater than sample size). The idea of using support vectors with discriminant analysis has been previously introduced in \cite{gu2010discriminant} which made significant improvement over LDA. In addition, the effectiveness of SVDA in i-Vector/PLDA speaker recognition for NIST SRE2010 is studied in \cite{fahimehICASSP} previously for both long and short duration test utterances.

More specifically, LDA defines speaker classes separation criterion in direction $A$ as,

       \begin{equation}
        \lambda=\dfrac{A^TS_bA}{A^TS_wA}, 
        \label{eq5}
      \end{equation}

where $S_b$ and $S_w$ are between class and within class covariance matrices. In traditional LDA every sample of all classes participate in calculating these covariance matrices; however, for SVDA only support vectors are used. The between class covariance matrix in SVDA is defined as,

       \begin{equation}
        S_b =\sum_{1 \le c_1 \le c_2 \le C}w_{c_1c_2}w_{c_1c_2}^T.
        \label{eq8}
      \end{equation} 
      
Where $w_{c_1c_2}$ is the optimal direction to separate two classes $c_1$ and $c_2$ by a linear SVM (for calculating $w_{c_1c_2}$ only support vectors of the two classes $c_1$ and $c_2$ are participating). If we define $\hat{X}=[\hat{x}_1,\hat{x}_2, ... , \hat{x}_{\hat{N}}]$ to contain all support vectors and $\hat{N}$ to be their total number; then, the within class covariance matrix for SVDA will be formulated as,

       \begin{equation}
        S_w =\sum_{c=1}^{C}\sum_{i \in \hat{I}_{c}}(\hat{x}_i - \hat{\mu}_c)(\hat{x}_i - \hat{\mu}_c)^T,
        \label{eq9}
      \end{equation} 
the index for support vectors in class $c$ and their mean are represented by $\hat{I}_{c}$ and $\hat{\mu}_c$, respectively. Finally, similar to LDA, the optimum transformation $\hat{A}$ will contain the $k$ eigenvectors corresponding to the $k$ largest eigenvalues of $S^{-1}_wS_b$. More details on the advantages and properties of SVDA are provided in \cite{fahimehICASSP}.

Two strategies can be adopted here for training linear SVM in SVDA framework: 1) one-versus-one and 2) one-versus-rest. We used the second approach as it can use minor and major unlabeled data optimally. More specifically, for training the SVM classifier to separate one class against data from the rest classes, minor and major unlabeled data are added to the rest class. Therefore, the class labels are not needed here.

\subsection{Unlabeled data PLDA}
To fully explore the information from the in-domain unlabeled data, we did an interesting experiment, which uses only the in-domain unlabeled data to train PLDA (however, SRE04-08 are used for training LDA). To do that, we use the ``estimated'' labels from speaker clustering. Surprisingly, PLDA with only 75 estimated speakers for 200 minor language i-Vectors achieved 20.5\% EER on Dev experiment (using i-Vectors for CRSS1 baseline), which is not so bad. However, if we add more data (i.e., 75 + 300 estimated speakers from 2472 i-Vectors in minor and major languages) for PLDA training, the performance degraded from 20.5\% to 26.3\%. This observation suggests that out-of-domain language data is not helpful to train a discriminative classifier, because in the view of Dev enrollment/test data, the major language data is also out-of-domain. We argue that even for data-driven algorithm such as PLDA, choosing a proper data set to train the classifier is still essential.

Motivated by this, we believe the use of unlabeled data in the SRE 2016 evaluation will be more beneficial. Compared with only 200 minor language utterances, major language set has 2272 utterances. Although the speaker label is not given, we say the estimated label is still useful, and could probably perform better than 20.5\% EER in the Eval set.

\subsection{Calibration and fusion}
The CRSS calibration and fusion system is mainly based on the BOSARIS toolkit \cite{brummer2013bosaris}. The PAV algorithm is used to create calibration transformation matrix. We used two data sets for calibration. The first one is Dev data, we use all the Dev trials information that NIST provided for system development to train the calibration system. Again, because in this SRE, the Dev and evaluation set have totally different languages, it's not guaranteed that the calibration will work well for the final Eval set. For this consideration, we created a new trial list to calibrate evaluation scores, and we used unlabeled data with estimated speaker labels. We believe the score distribution of unlabeled data will be closer to that of evaluation.

After calibration, we fused our sub-systems for final submission. For system fusion, we employed a simple linear fusion system using logistic regression.

\section{CRSS sub-systems}
\label{sec:exp}
We developed 7 sub-systems from 4 CRSS baselines that used SREs to train SVDA, LDA and PLDA. Also, as described above, we developed 4 sub-systems with just unlabeled data PLDA idea. The details of each system about data and techniques they used are listed in Table \ref{Table1} and Table \ref{Table2}. More specifically, sub-systems 8, 9, 10, 11 are respectively share the same configuration as sub-systems 3, 4, 6, 2; however, only unlabeled data are used to train PLDA.

\begin{table}[!th]
	\centering
	\caption{\label{Table2} {\it Description of CRSS sub-systems using just unlabeled data to train PLDA.}}
	\vspace{2mm}
		\begin{tabular}{|c|c|c|c|}
			\hline
		Sub-system & i-Vector & SVDA  & LDA   \\  
			\hline\hline
			 8 & CRSS 1  & \xmark &\cmark\\
			\hline
			 9 & CRSS 2 &\xmark & \cmark \\
			\hline
			 10 & CRSS 3  &\cmark&\cmark \\
			\hline
			 11 & CRSS 4 &\cmark &\cmark\\
			\hline		
		\end{tabular}
\end{table}

\begin{table}[h]
  \centering
  \caption{\label{tab-2}{\it CRSS submission for NIST SRE2016.}}
  \vspace{1mm}
  \begin{tabular}{|c| c | c | c | }
  \hline
   Submission & sub-systems & Calibration Data  & Fusion   \\
 \hline
 \hline
  Primary & 1-7 & Dev+Unlabeled & LR  \\
   Contrastive1 & 1-7 & Dev & LR   \\
  Contrastive2 & 1-11  & Dev+Unlabeled &  LR \\ 
  \hline
  \end{tabular}

  \end{table}

      \begin{table*}[!th]
        \caption{\label{tab:single_sys_score} {\it Calculated scores for the single systems. With using different data for calibration just the value of act-Cprimary will be changed. These scores are calculated using NIST scoring software for the Dev set, the equalized and unequalized scores are separated with -- in the table.}}
        \vspace{2mm}
        \centering
        \resizebox{15cm}{!}{
		\begin{tabular}{ |c||c||c|c|c|c| }
  			\hline
			Sub &  EER/min-Cprimary &  \multicolumn{3}{c |}{act-Cprimary} \\  
  			system &  & Calibrate With Dev & Calibrate With Unlabeled & Calibrate With Dev+Unlabeled \\
  			\hline
  			\hline
  			1 & 17.14/0.768 -- 18.64/0.779 &  0.768 -- 0.779 &  0.881 -- 0.891 & 0.812 -- 0.822\\
  			2 & 17.11/0.755 -- 18.87/0.757 & 0.768 -- 0.769 & 0.794 -- 0.8 & 0.777 -- 0.775\\
  			3 & 17.84/0.754 -- 18.41/0.734 & 0.754 -- 0.734 & 0.834 -- 0.826 & 0.788 -- 0.766\\
  			4 & 17.17/0.719 -- 17.50/0.694 & 0.722 -- 0.694 & 0.820 -- 0.813 & 0.769 -- 0.747\\
  			5 & 15.59/0.701 -- 16.08/0.671 & 0.709 -- 0.671 & 0.813 -- 0.813 & 0.747 -- 0.726\\
  			6 & 15.58/0.679 -- 15.95/0.629 & 0.688 -- 0.629 & 0.744 -- 0.735 & 0.694 -- 0.647\\
  			7 & 15.53/0.685 -- 16.63/0.658 & 0.686 -- 0.658 & 0.775 --  0.769 & 0.697 -- 0.672\\
  			\hline
		\end{tabular}
		}
      \end{table*}  
      
\section{CRSS submissions}
\label{submissions}
The final submissions of CRSS is the fusion of several sub-systems. In the final submission, we tried different system combinations as well as different calibration strategies. We submit a 1-7 sub-systems fusion with Dev+unlabeled data for calibration as our primary submission. To test our hypothesis that unlabeled PLDA idea will benefit for the Eval set, we submitted this as a contrastive submission. All these combinations make our final submissions to SRE 2016.

\section{Performance of CRSS sub-systems on SRE 2016 Dev data}
\label{performance}
Tables \ref{tab:single_sys_score}, \ref{unlabPldaScores}, and \ref{fusionScore}  shows the equal error rate (EER), minimum Cprimary(min-Cprimary) and actual Cprimary (act-Cprimary) costs for single systems and fusion systems using NIST scoring software. These results are evaluated on Dev set.

  \begin{table}[!th]
\caption{{\it EER, min-Cprimary and act-Cprimary costs for single systems that only use unlabeled data for training PLDA. NIST scoring software is used to calculate the scores for Dev set. The equalized and unequalized scores are separated with -- in the table. For the calibration Dev + unlabeled data are used.}}
\vspace{2mm}
	\centering
		\resizebox{7cm}{!}{
		\begin{tabular}{|c|c|c|}
			\hline
		Sub & EER/min-Cprimary & act-Cprimary   \\  
		system &  &    \\ 
			\hline\hline
			 8 &29.48/0.901 -- 26.84/0.9 & 0.917 -- 0.914 \\
			\hline
			 9 & 29.72/0.898 -- 26.15/0.908 & 0.92 -- 0.933 \\
			\hline
			 10 & 26.48/0.943 -- 24.96/0.954  & 0.956 -- 0.96\\
			\hline
			 11 & 26.76/0.957 -- 26.45/0.968 & 0.986 -- 0.989\\
			\hline		
		\end{tabular}
	}
	\label{unlabPldaScores} 
\end{table}

\begin{table}[h]

\caption{{ \it Fusion scores calculated on Dev set with NIST scoring software. The equalized and unequalized scores are separated with -- in the table.}}
\vspace{2mm}
  \centering
  \resizebox{8cm}{!}{
  \begin{tabular}{|c| c | c | }
  \hline
   Submission & EER/minCprimary & act-Cprimary   \\
 \hline
 \hline
  Primary & 14.24/0.590 -- 14.98/0.561 & 0.612 -- 0.58 \\
  Contrastive1 & 13.81/0.585 -- 14.66/0.554 & 0.589 -- 0.56 \\
  Contrastive2 & 14.27/0.592 -- 15.04/0.562 & 0.618 -- 0.589 \\ 
  \hline
  \end{tabular}
  }
  \label{fusionScore}
  \end{table}
      
\section{COMPUTATIONAL RESOURCES}
\label{Computational}
\subsection{CPU cluster}
The speaker recognition system was implemented on our in-house high-performance Dell computing cluster, running Rocks 6.0 (Mamba) Linux distribution. The cluster comprises of eight 6C Intel Xeon 2.67 GHz CPU’s, four 10C Intel Xeon 2.40 GHz CPU’s, and 18 quad-core Intel Xeon 2.33 GHz CPU’s, yielding a total of 408 processors. The total amount of internal RAM on the cluster exceeds 1 TB. All our data including audio files, features, statistics, etc. are stored on a 30 TB Dell PowerVault MD1000 direct attached storage.
\subsection{GPU machines}
For DNN training on SWB data, GeForce GTX TITAN Black graphic card is used, 6144 MB Ram. For DNN training on Fisher English, we used a 12 GB Tesla K40.

\subsection{CPU execution time}
We tested the system’s scoring process using one CPU
of 2.67 GHz clock speed and 32 GB RAM. We selected a
3 minute utterance (exact duration of 181.45 seconds) and calculated the time required to perform feature extraction (20 dimensional MFCC), voice activity detection (Kaldi SRE10/v1 default), extraction of zero and first order statistics and the 600 dimensional i-Vector. The time required for this chain of processes is for the selected utterance is 37.58s. This is computed by averaging the elapsed time obtained from three independent runs. Scoring an utterance using our PLDA model takes less than 0.1 seconds on average. For training the models, it depends on how many enrollment utterances are provided. Since the UBM and TV matrices are trained off-line, speaker enrollment requires only to extract the corresponding i-Vectors, thus the time required will be a multiple of the number of enrollment utterances provided for a speaker.

\section{Acknowledgement}
We would like to thank Dr. Kong Aik Lee for providing the unlabeled trial list, organizing I4U meetings, and other I4U group members sharing many ideas and insights during I4U meetings. We would like to thank Qian Zhang, Abhinav Misra, Dr. Finnian Kelly and other CRSS colleagues for their many insights and helpful discussions in the development of the systems. We would like to thank Dr. Gang Liu (CRSS graduate) from Alibaba Group for providing the fusion system. 

\vfill
\pagebreak

\urlstyle{same}
\bibliographystyle{IEEEbib}
\bibliography{reference}

\end{document}